\documentclass[twoside,11pt]{article}
\usepackage[abbrvbib, preprint]{jmlr2e}
\usepackage{array,graphicx}
\usepackage{booktabs}
\usepackage{multirow}
\usepackage{pifont}
\usepackage{caption}
\usepackage{subcaption}
\usepackage{pythonhighlight}
\usepackage{listings}

\lstdefinestyle{Python}{
    language        = Python,
    basicstyle      = \ttfamily,
    keywordstyle    = \color{blue},
    keywordstyle    = [2] \color{teal}, % just to check that it works
    stringstyle     = \color{green},
    commentstyle    = \color{red}\ttfamily
}

\usepackage{cleveref}
\definecolor{codegreen}{rgb}{0,0.6,0}
\definecolor{codegray}{rgb}{0.5,0.5,0.5}
\definecolor{codepurple}{rgb}{0.58,0,0.82}
\definecolor{backcolour}{rgb}{0.95,0.95,0.92}
\newcommand*\rot{\rotatebox{90}}

\begin{document}
%%% Mehrere Autoren werden durch \and voneinander getrennt.
%%% Die Fußnote enthält die Adresse sowie eine E-Mail-Adresse.
%%% Das optionale Argument (sofern angegeben) wird für die Kopfzeile verwendet.
\title{TSInterpret: A unified framework for time series interpretability}
%%%\subtitle{Untertitel / Subtitle} % if needed
\author{\name Jacqueline Höllig \email hoellig@fzi.de\\
       \addr Information Process Engineering\\
       FZI Forschungszentrum Informatik\\
       76131 Karlsruhe, Germany
       \AND
       \name Cedric Kulbach \email kulbach@fzi.de\\
       \addr Information Process Engineering\\
       FZI Forschungszentrum Informatik\\
       76131 Karlsruhe, Germany
       \AND
       \name Steffen Thoma \email thoma@fzi.de\\
       \addr Information Process Engineering\\
       FZI Forschungszentrum Informatik\\
       76131 Karlsruhe, Germany     }

%\editor{Kevin Murphy and Bernhard Sch{\"o}lkopf}

\maketitle

\begin{abstract}
With the increasing application of deep learning algorithms to time series classification, especially in high-stake scenarios, the relevance of interpreting those algorithms becomes key. 
Although research in time series interpretability has grown, accessibility for practitioners is still an obstacle. Interpretability approaches and their visualizations are diverse in use without a unified api or framework. To close this gap, we introduce TSInterpret\footnote{\url{https://github.com/fzi-forschungszentrum-informatik/TSInterpret}}, an easily extensible open-source Python library for interpreting predictions of time series classifiers that combines existing interpretation approaches into one unified framework. 
The library features (i) state-of-the-art interpretability algorithms, exposes a (ii) unified API enabling users to work with explanations in a consistent way, and provides (iii) suitable visualizations for each explanation.
\end{abstract}
%%% Beginn des Artikeltexts
\section{Introduction}
%The increasing deployment of artificial intelligence (AI) systems in
%high stakes domains has been coupled with an increase in societal
%demands for these systems to provide explanations for their predictions. However, many machine %learning techniques are not easily
%explainable, even by experts in the field. This has led to a growing
%research community, with a long history, focusing on “interpretable”
%or “explainable” machine learning techniques. However, despite the
%growing volume of publications, there remains a gap between what
%%society needs and what the research community is producing. One
%reason for this gap is a lack of a precise definition of an explanation
%as different people in different settings may require different kinds
%of explanations. For example, a doctor trying to understand an AI
%diagnosis of a patient may benefit from seeing known similar cases
%with the same diagnosis; a denied loan applicant will want to understand the main reasons for their rejection and what can be done
%to reverse the decision; a regulator, on the other hand, will want to
%understand the behavior of the system as a whole to ensure that it
%complies with the law; and a developer may want to understand
%where the model is more or less confident as a means of improving
%its performance.
%With the increasing accuracy of deep learning models, the application on time series classification increased.
%The rising exertion of deep learning methods in safety-critical scenarios leads to a higher need for interpretability. 
Although time series data are omnipresent in industry and daily life, deep learning methods have only been applied to time series data in the last decade. Before, time series classification (TSC) primarily focused on classifications based on discriminatory features (e.g., learning based on the whole time series, intervals, or shapelets (\cite{bagnall_great_2017})). Compared to deep methods, they do not have the  ”black-box”- character as a significant drawback. However, with the increasing accuracy of deep learning models and their ability to cope with vast amounts of data, their application on time series classification increased, leading to a need for interpretability, especially in high-risk settings.
% like disease prediction or autonomous driving. %the ”black-box”- character of such models is a significant drawback. In such settings, the interpretability of the decision process becomes crucial. 
Various interpretability methods for time series classification are available (\cite{rojat_explainable_2021}). However, the usage of those methods is not yet standardized: The proposed methods often lack a) open code (e.g., \cite{siddiqui_tsinsight_2021}), b) an easy-to-use interface (e.g., \cite{ismail_benchmarking_2020}), or c) visualization (e.g., \cite{guilleme_agnostic_2019}), making the application of those methods inconvenient and thereby hindering the usage of deep learning methods on safety-critical scenarios. \par
Although unifying frameworks (e.g., tf-explain (\cite{meudec_raphael_tf-explain_2021}), Alibi (\cite{klaise_alibi_2021}), or captum (\cite{kokhlikyan_captum_2020})) have already been developed for various data types (tabular, image, text) and algorithms (e.g., GradCam (\cite{selvaraju_grad-cam_2020}), Counterfactuals (\cite{mothilal_explaining_2020}), SHAP (\cite{lundberg_unified_2017})), an interpretability framework for time-series data is still missing. Due to the different structures and properties of time-ordered data, most approaches and frameworks are not directly applicable to time series (\cite{ismail_benchmarking_2020}). Further, time series are not intuitive for human understanding (\cite{siddiqui_tsviz_2019}) and need additional visualizations.
Therefore, we propose TSInterpret, a framework implementing algorithms for time series classification. In this work, we provide
\begin{itemize}
    \item a review of existing interpretability libraries,
    \item a unified framework for time series interpretability,
    \item and unified visualizations for the implemented algorithms.
\end{itemize}
%The increasing popularity and preformance of machine learning models on time series classification, has lead to adaption into high staked domains. These Adaptions have been coupled with an increase of societal demands to provide an explanations for these systems. \par

\section{Need for Interpretability in Time Series Classification}\label{sec:Motivation}
Temporal data is ubiquitous and encountered in many real-world applications ranging from electronic health records (\cite{rajkomar_scalable_2018}) to cyber security (\cite{susto_time-series_2018}). Although almost omnipresent, time series classification has been considered one of the most challenging problems in data mining for the last two decades (\cite{yang_10_2006}; \cite{esling_time-series_2012}). With the rising data availability and accessibility (e.g., provided by the UCR / UEA archive (\cite{bagnall_great_2017, dau_ucr_2019})), hundreds of time series classification algorithms have been proposed. Although deep learning methods have been successful in the field of Computer Vision (CV) and Natural Language Processing (NLP) for almost a decade, application on time series has only occurred in the past few years (e.g., \cite{fawaz_deep_2019, rajkomar_scalable_2018,susto_time-series_2018,ruiz_great_2021}). %The movement from classifications based on hand-crafted features and statistics to deep learning methods led to a need for time series interpretability. 
Deep learning models have been shown to achieve state-of-art results on time series classification (e.g., \cite{fawaz_deep_2019}). However, those methods are black-boxes due to their complexity which limits their application to high-stake scenarios (e.g., in medicine or autonomous driving), where user trust and understandability of the decision process are crucial.
\par
Although much work has been done on interpretability in CV and NLP, most developed approaches are not directly applicable to time series data. The time component impedes the usage of existing methods  (\cite{ismail_benchmarking_2020}). Thus, increasing effort is put into adapting existing methods to time series (e.g., LEFTIST based on SHAP / Lime (\cite{guilleme_agnostic_2019}), Temporal Saliency Rescaling for Saliency Methods (\cite{ismail_benchmarking_2020}), Counterfactuals (\cite{ates_counterfactual_2021, sanchez-ruiz_instance-based_2021})), and developing new methods specifically for time series interpretability (e.g., TSInsight based on autoencoders (\cite{siddiqui_tsinsight_2021}), TSViz for interpreting CNN (\cite{siddiqui_tsviz_2019})). For a survey of time series interpretability, please refer to \cite{rojat_explainable_2021}.
\par
Compared to images or textual data, humans cannot intuitively and instinctively understand the underlying information contained in time series data. Therefore, time series data, both uni- and multivariate, have an unintuitive nature,  lacking an understanding at first sight (\cite{siddiqui_tsviz_2019}). Hence, providing suitable visualizations of time series interpretability becomes crucial. \par

For instance, health data tasks like cardiac disease detection from electrocardiogram data (ECG5000, \cite{dau_ucr_2019}) or seizure prediction from the data of a tri-axial accelerometer (Epilepsy, \cite{bagnall_uea_2018}) are examples for the need of interpretability within time series data. Those classification tasks are in a sensitive field regarding people’s lives. Therefore, decisions need to be carefully taken based on solid evidence. Wrong medications or unidentified diseases can have long-lasting effects on a patient's health. To allow physicians to make such data-driven decisions with the help of machine learning, interpretations of black-box models are crucial. It is not only relevant if a patient has epilepsy or a cardiac disease but why or which indications are there for such. Further, implementing a non-interpretable machine learning model in medicine raises legal and ethical issues (\cite{amann_explainability_2020}).
\section{Related Work}\label{sec:RelatedWork}
\begin{table}[tbh]
    \centering
    \resizebox{\textwidth}{!}{%
    \begin{tabular}{|c|c|c|c|c|c|c|c|c|c|c|c|c|c|c|}
    \hline
    \multirow{2}{*}{Name} & \multirow{2}{*}{Backend} & \multicolumn{4}{c}{Data}&\multicolumn{2}{|c|}{Task}&\multicolumn{4}{c|}{Scope}&\multicolumn{3}{c|}{Output}\\
     \cline{3-15}
    & &\rot{Tabular}& \rot{Text}& \rot{Image}& \rot{Time Series}&\rot{Classification} &\rot{Regression} & \rot{Model Agnostic} & \rot{Model Specific} & \rot{Local} & \rot{Global} &\rot{FA} & \rot{IB} & \rot{RB}\\
    
    \hline
    AIX360 (\cite{arya_one_2019}) & PYT,TF, BB & X&X&X& & X&X & X &X&X&X&X&X&X\\
    \hline
    ALIBI (\cite{klaise_alibi_2021}) & BB &X&X&X& & X& X  &X&X&X&X&X&X&\\
    \hline
    Anchor (\cite{ribeiro_anchors_2018}) & BB & X&X&&&&&X&&X&&&X&\\
    \hline
    Captum (\cite{kokhlikyan_captum_2020}) &PYT &X& X& X& & X&X & X&X&X&&X&&\\
    \hline
    carla (\cite{pawelczyk_carla_2021}) & PYT, TF, SK & X&&&& X&X & X&&X&&&X&\\
    \hline
    DALEX (\cite{baniecki_dalex_2021}) & SK,TF & X&&&& X&X&X&&X&X&X&&\\
    \hline
    DeepExplain (\cite{ancona_towards_2018}) & TF & &X&X&& X&X&  & & X&X&X&&\\
    \hline 
    Dice  (\cite{mothilal_explaining_2020}) & PYT, TF, BB & X&&&& X&X  & X&&X&&&X&\\
    \hline
    ELI5 (\cite{eli5-org_eli5_nodate}) & BB & X&X&X&&X& & X& &X&&X&&\\
    \hline
    %Ethical ML \\
    %\hline 
    H2O (\cite{hall_machine_nodate})& - &X&&&X& X&X&X&X&X&X&X&&\\
    \hline
    iNNvestigate (\cite{alber_innvestigate_2019})&TF& &&X& &X&  & &X&X&X&X&&\\
    \hline
    InterpretMl (\cite{nori_interpretml_2019}) & BB, WB  & X&&& &X&X &X&X&X&X&X&&\\
    \hline
    Lucid (\cite{tensorflow_lucid_nodate}) & TF & &&X& & X&X &&X&X&X&X&&\\
    \hline
    OmniXAI (\cite{yang_omnixai_2022}) & PYT, TF, SK &X &X&X&X & X&X &X&X&X&X&X&X&\\ \hline
    pytorch-cnn-visualizations (\cite{ozbulak_pytorch_2019})& PYT &&&X& & X& &&X&X&&X&&\\
    \hline
    SHAP (\cite{lundberg_unified_2017}) & PYT, TF, BB & X&X&X& &X&X&X&X&X&X&X&&\\
    \hline
    Skater (\cite{oracle_skater_nodate})& BB & X& X& X&& X&X & &&X&X&X&X&\\
    \hline
    Tf-Explain (\cite{meudec_raphael_tf-explain_2021}) & TF& &&X& &X& &&X&X&&X&&\\
    \hline
    TorchRay (\cite{fong_understanding_2019}) & PYT &&&X&&X& &&X&&&X&&\\
    \hline
    What-if (\cite{wexler_what-if_2019}) & TF&X&X&X&& X&X&X&&&&&X&\\
    \hline
    wildboar (\cite{samsten_isaksamstenwildboar_2020})&-&&&& X & X&X&&X&X&&&X&\\
    \hline
    \end{tabular}}
    \caption{Overview over recent explanation libraries and their time series capabilities. \texttt{BB}: Black-Box, \texttt{TF}: Tensorflow, \texttt{PYT}: PyTorch, \texttt{SK}: scikit-learn}
    \label{tab:my_label}
\end{table}
Numerous open-source tools for machine learning evolved in the last few years. In the following, we focus on libraries implementing various post-hoc interpretability methods installable via PyPI. For the summary in \Cref{tab:TSInterpret}, only active libraries, i.e., libraries with some development activity in the past 12 months on GitHub, are included. The comparison is conducted according to the scope of included explanations (Model Agnostic vs. Model-specific, Global vs. Local) and library properties (supported model libraries, data types, and tasks). Model Agnostic methods apply to any classification algorithms, whereas model-specific algorithms only work on a subset of classification algorithms (e.g., Cam on Convolutional Neural Networks). Global interpretability methods interpret the model's overall decision-making process, while local interpretability focuses on the interpretation of a single instance and the prediction of that instance\footnote{More information on the taxonomy are available in \cite{rojat_explainable_2021}.}. Further, the return type of the explanation methods are taken into account. Feature Attribution methods (FA) return a per-feature attribution score based on the feature’s contribution to the model’s output (e.g., GradCam (\cite{selvaraju_grad-cam_2020}) or SHAP (\cite{lundberg_unified_2017})), instance-based methods (IB) calculate a subset of relevant features that must be present to retain or remove a change in the prediction of a given model (e.g., counterfactuals (\cite{mothilal_explaining_2020}) or anchors (\cite{ribeiro_anchors_2018})), while rule-based methods (RB) derive rules. \par

Except for the libraries wildboar (\cite{samsten_isaksamstenwildboar_2020}), OmniXAI (\cite{yang_omnixai_2022}), and H2O (\cite{hall_machine_nodate}), no library provides interpretability methods on time series. However, the support of OmniXAI (\cite{yang_omnixai_2022}) is restricted to anomaly detection, univariate time series and includes only non-time-series-specific interpretability methods. Wildboar (\cite{samsten_isaksamstenwildboar_2020}) focuses on temporal machine learning (classification, regression, and anomaly detection) and only provides a limited number of (counterfactual) interpretability methods as additional features.  A unified framework for time series interpretability like captum or ALIBI enabling easy-to-use interpretability is still missing.
\section{Library Design}
% begin descriptiom 
The API Design is orientated on the scikit design paradigms consistency, sensible defaults, composition, nonprofilarity of  classes, and inspection (\cite{buitinck_api_2013}). %\Cref{fig:Architecture} shows the structure of TSInterpret. 
\begin{figure}[!ht]
      \centering
      \includegraphics[width=.9\textwidth]{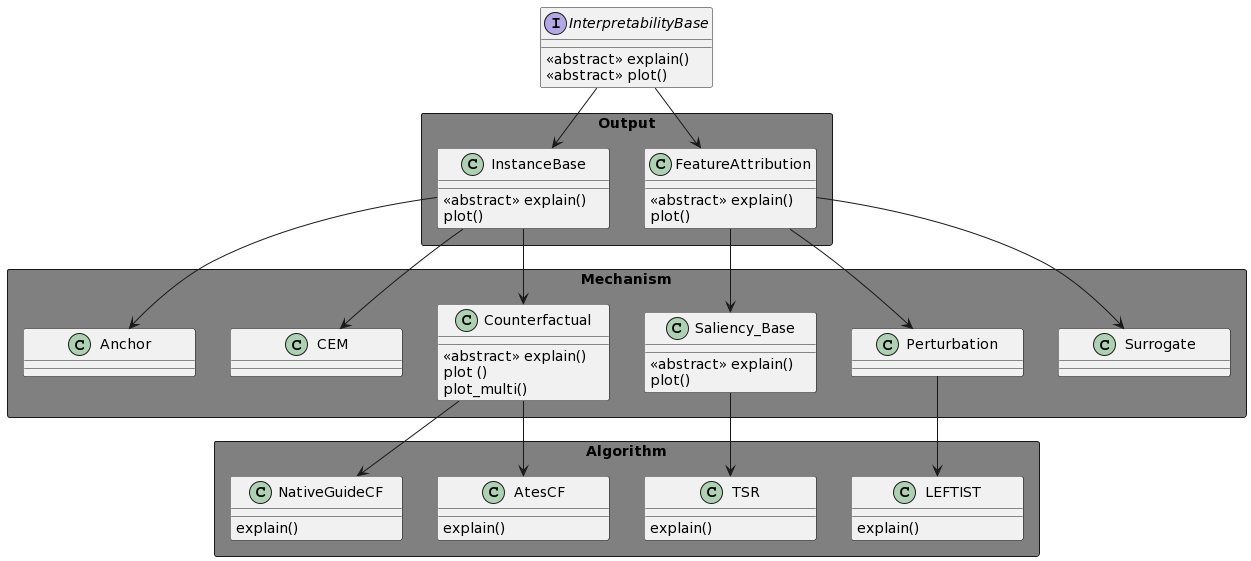}
      \caption{Structure of TSInterpret. }
      \label{fig:Architecture}
\end{figure}
\Cref{lst:InterpretabilityExample} shows the workflow of the library in a coding sample. Given a trained machine learning model and an instance to be classified: First, the desired interpretability method is imported (line 1) and instantiated (line 2), followed by explaining the instance (line 3), and finally, the generation of the plot (line 4).
\begin{lstlisting}[style=mypython,caption= API Usage Example., label=lst:InterpretabilityExample]
from TSInterpret.InterpretabilityModels.Saliency.TSR import TSR
    
int_mod = TSR(model, item.shape[-1], item.shape[-2])
exp = int_mod.explain(item, labels = label, TSR = True)
int_mod.plot(item,exp)
\end{lstlisting}
\begin{description}
  \item[Consistency] All implemented objects share a consistent interface. 
  %\Cref{fig:Architecture} shows the structure of the interfaces and their implementation. 
  Every interpretability method inherits from the interface \texttt{InterpretabilityBase} to ensure that all methods contain a method \texttt{explain} and a \texttt{plot} function. 
  The \texttt{plot} function is implemented on the level below based on the output structure provided by the interpretability algorithm to provide a unified visualization experience (e.g., in the case of Feature Attribution, the plot function visualizes a heatmap on the original sample). If necessary, those plots are refined by the Mechanism layer. This is necessary to ensure suitable representation as the default visualization can sometimes be misinterpreted (e.g., the heatmap used in the plot function of \texttt{InterpretabilityBase} allows positive and negative values, while  \texttt{TSR} is scaled to $[0,1]$. Using the same color pattern for both scales would lead to a high risk of misinterpreting results while comparing \texttt{TSR} with \texttt{LEFTIST}.). The \texttt{explain} function is implemented on the method level. 
  \item[Sensible Defaults] TSInterpret provides reasonable defaults for most parameters by providing the default parameterizations for each method from the designated papers. Those parameters can easily be changed if needed by providing alternative values during model instantiation.
  %\begin{lstlisting}[language=Python, breaklines=true, caption=Example Signature of the method explain for LEFTIST. In the signature the sensible defaults are set., label=lst:Hyperparameter]
  %  def explain(self,instance,nb_neighbors, idx_label=None, explanation_size=None,transform_name='straight', segmentator_name='uniform', learning_process_name='Lime',nb_interpretable_feature=10, random_state=0):
  %\end{lstlisting}
  \item[Composition] Many interpretability methods for time series classification are based on already existing methods for tabular, image, or text data (e.g., \cite{ismail_benchmarking_2020} is based amongst others on \cite{lundberg_unified_2017}). Whenever feasible, existing implementation of such algorithms are used (e.g., SHAP (\cite{lundberg_unified_2017}), captum (\cite{kokhlikyan_captum_2020}), or tf-explain (\cite{meudec_raphael_tf-explain_2021}).
  \item[Nonprofilarity of classes] \hfill \\ TSInterpret implements the interpretability algorithms as custom classes. Datasets, instances, and results are represented as NumPy arrays, Lists, or Tuples, instead of classes. For instance, the counterfactual method returns a Tuple of the counterfactual time series and the label (list, int). Hyperparameters are regular strings or numbers. 
  
  %  \begin{lstlisting}[language=Python, breaklines=true, caption=Format of Counterfactual return., label=lst:Return]
  %  Tuple(counterfactual, label)\end{lstlisting}
  \item[Inspection] TSInterpret stores and exposes the parameter of the interpretability algorithms as public attributes. In some methods, parameters have a significant impact on the obtained results. Making those parameters publicly available through attributes facilitates experimenting with hyperparameters.
  
\end{description}

\section{Algorithm Overview}
As depicted in \cref{sec:RelatedWork}, our framework provides support for various time series interpretability types, resulting in interpretations and visualization for a variety of use cases. The current version of the library includes the interpretability algorithms listed in \Cref{tab:TSInterpret}. Their implementation in TSInterpret is based on code provided by the authors of the algorithms, which got adapted to the framework and extended to ensure the availability on multiple model backends (e.g., PyTorch or TensorFlow).
\begin{table}[!bth]
    \centering
   
    \begin{tabular}{|l|cccc|}
        \hline
         \textbf{Method} & \textbf{Model} & \textbf{Explanations} & \textbf{Type} & \textbf{Dataset}   \\ \hline
         NUN-CF (\cite{sanchez-ruiz_instance-based_2021}) & TF, PYT, SK & IB & uni & y\\
         CoMTE (\cite{ates_counterfactual_2021}) &  TF, PYT,SK & IB &multi&y\\ 
         LEFTIST (\cite{guilleme_agnostic_2019}) &TF, PYT, SK & FA &uni&y\\ 
         %SHAP Time &  &&&&\\
         TSR (\cite{ismail_benchmarking_2020}) & TF, PYT & FA&multi& n\\ \hline
         %TS Insight & &&&&\\
         %END MINIMAL SCOPE & &&&&\\
    \end{tabular}
    \caption{Interpretability Methods implemented in TSInterpret. \texttt{BB}: Black-Box, \texttt{TF}: Tensorflow, \texttt{PYT}: PyTorch, \texttt{SK}: scikit-learn}
    \label{tab:TSInterpret}
\end{table}
\begin{description}

\item[NUN-CF]\footnote{\url{https://github.com/e-delaney/Instance-Based_CFE_TSC}} \cite{sanchez-ruiz_instance-based_2021} proposed using the K-nearest neighbors from the dataset belonging to a different class as native guide to generate counterfactuals. They propose three options for transforming the original time series with this native guide: the plain native guide, the native guide with bary centering, and transformation based on the native guide and class activation mapping.
\item[CoMTE]\footnote{\url{https://github.com/peaclab/CoMTE}} \cite{ates_counterfactual_2021} proposed CoMTE as a perturbation-based approach for multivariate time series counterfactuals. The goal is to exchange the smallest possible number of features with reference data by applying random-restart hill climbing to obtain a different classification.\\
\item[LEFTIST]\footnote{\url{https://www.dropbox.com/s/y1xq5bhpf0irg2h/code_LEFTIST.zip?dl=0}} Agnostic Local Explanation for Time Series Classification by \cite{guilleme_agnostic_2019} adapted LIME for time series classification and proposed to use prefixed shapelets as the interpretable components. Each shapelet is a non-overlapping subsection of the original time series with a prefixed length. The feature importance is provided on a shapelet basis.
\item[TSR]\footnote{\url{https://github.com/ayaabdelsalam91/TS-Interpretability-Benchmark}} Temporal Saliency Rescaling (\cite{ismail_benchmarking_2020}) calculates the importance of each timestep, followed by the feature importance based on different Saliency Methods, both back-propagation based and perturbation based. We refer the reader to our code documentation for a complete list of implemented methods. The implementation in TSInterpret is based on tf-explain (\cite{meudec_raphael_tf-explain_2021}), SHAP (\cite{lundberg_unified_2017}) and captum (\cite{kokhlikyan_captum_2020}).
\end{description}

%\begin{figure}[!h]
%     \centering
%     \begin{subfigure}[b]{0.4\textwidth}
%         \centering
%         \includegraphics[width=\textwidth]{figures/Nun_Cf.png}
%         \caption{Counterfactual created with NUN-CF on the Electric Devices Dataset.}
%         \label{fig:Nun-Cf}
%     \end{subfigure}
%     \hfill
%     \begin{subfigure}[t]{0.4\textwidth}
%         \centering
%         \includegraphics[width=\textwidth]{figures/Leftist.png}
%         \caption{Visualization of feature attributes obtained %with LEFTIST.}
%         \label{fig:Leftist}
%     \end{subfigure} \newline
%    \begin{subfigure}[b]{0.4\textwidth}
%%         \centering
%        \includegraphics[width=\textwidth]{figures/Ates.png}
%         \caption{Vizualization of the counterfactuals Feature obtainted by CoMTE on NATOPS.}
%         \label{fig:ates}
%     \end{subfigure}
%     \hfill
%     \begin{subfigure}[b]{0.4\textwidth}
%         \centering
%         \includegraphics[width=\textwidth]{figures/Ismail.png}
 %        \caption{Saliency obtained by TSR on Basic Motion.}
 %        \label{fig:Ismail}
 %    \end{subfigure}
 %     \caption{Default Visualizations obtained by TSInterpret. }
 %       \label{fig:Vis}
%\end{figure}

\begin{figure}[t]
\centering
     \begin{subfigure}[t]{0.45\textwidth}
         \includegraphics[width=\textwidth]{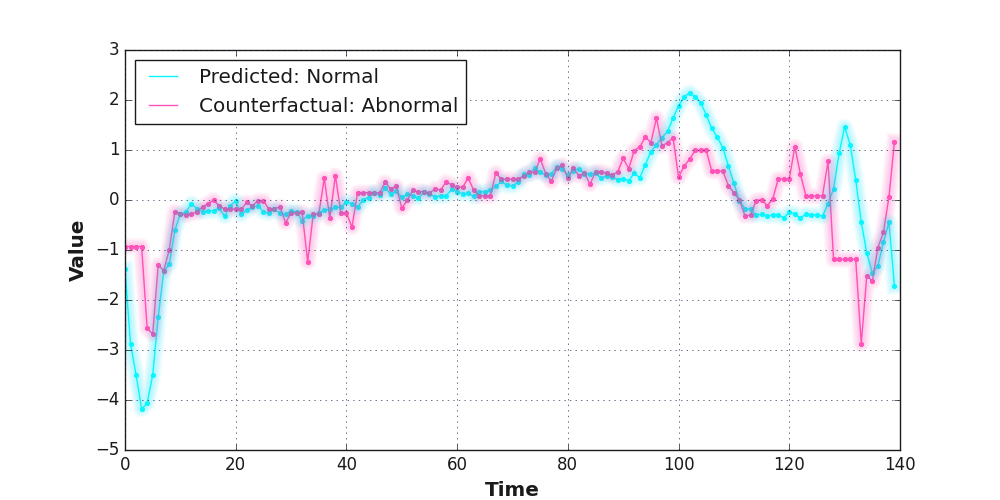}
         \caption{NUN-CF.}
         \label{fig:Nun-Cf}
     \end{subfigure}
      \begin{subfigure}[t]{0.3\textwidth}
         \includegraphics[width=\textwidth]{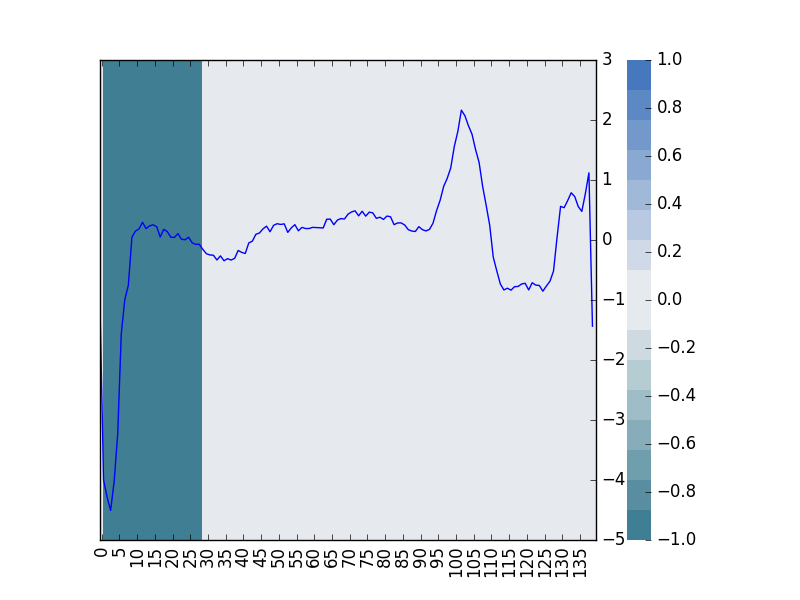}
         \caption{LEFTIST.}
         \label{fig:Leftist}
     \end{subfigure}
\caption{Interpretations obtained with TSInterpret the univariate ECG5000 dataset.}
\label{fig:Vis_univariate}     
\end{figure}

\begin{figure}[b]
     \centering
     \begin{subfigure}[t]{0.5\textwidth}
         \centering
         \includegraphics[width=\textwidth]{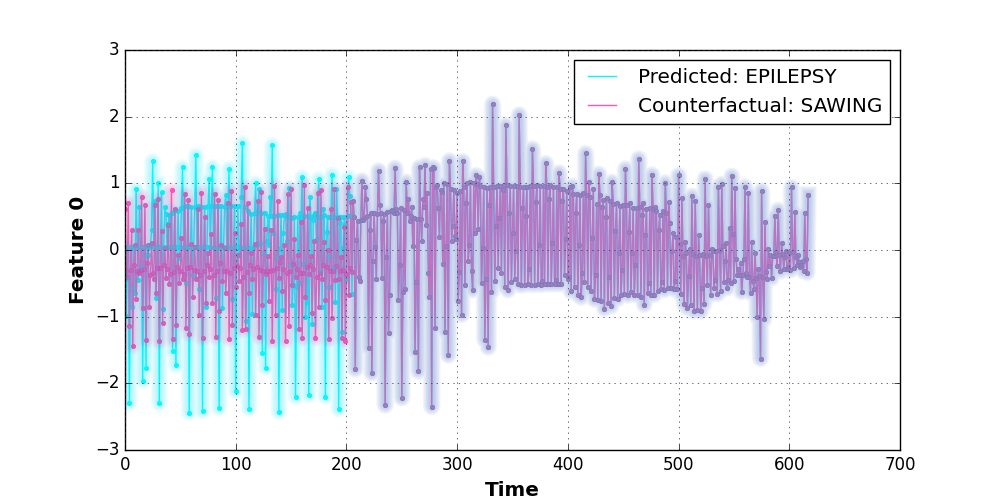}
         \caption{CoMTE.}
         \label{fig:ates}
     \end{subfigure}
      \begin{subfigure}[t]{0.35\textwidth}
         \centering
         \includegraphics[width=\textwidth]{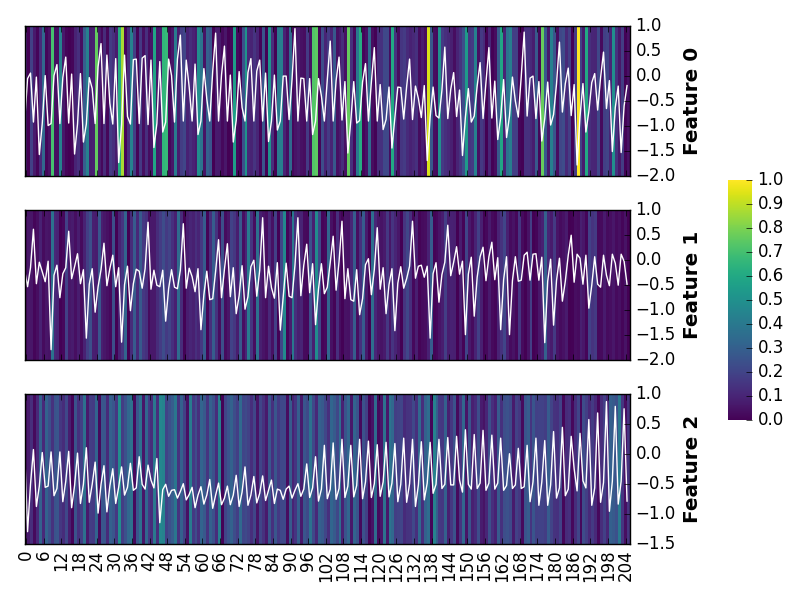}
         \caption{TSR.}
         \label{fig:Ismail}
     \end{subfigure}
\caption{Interpretations obtained with TSInterpret the multivariate Epilepsy dataset.}
\label{fig:Vis_multivariate}     
\end{figure}
\par
Returning to the use case described in \cref{sec:Motivation}, we train a 1D-conv ResNet on ECG5000 and Epilepsy. \Cref{fig:Vis_univariate} and \Cref{fig:Vis_multivariate} show interpretations obtained with TSInterpret for the 1D-conv ResNet on the first item of the testset for the univariate ECG5000 dataset (\cref{fig:Vis_univariate}) and the multivariate Epileptical dataset (\cref{fig:Vis_multivariate}). With the help of the interpretations shown in \Cref{fig:Vis_univariate} a physician can gather evidence on the functionality of the ResNet classification on ECG 5000. The feature attribution method LEFTIST (\cref{fig:Leftist}) identifies sections with positive/ negative influence on the current time series prediction. In this case, the first section of the time series has the most considerable impact on the current prediction being a normal electrocardiogram. If a physician wants to know why the electrocardiogram shows a normal instance and not an abnormal beat, conclusions can be drawn from the counterfactual approach NUN-CF. Compared to the original time series (in blue \cref{fig:Nun-Cf}) classified as normal, the classification is changed if the drop at the beginning is lower and the drop at the end is replaced with a rise (pink line in \cref{fig:Nun-Cf}). \Cref{fig:Vis_multivariate} shows interpretations for the epilepsy dataset. Suppose a physician wants to know how the original instance is predicted as walking (blue) instead of running. In such a case, the practitioner applies CoMTE to generate a counterfactual in the class direction of running, resulting in a different acceleration of the x-axis in the first 200 time steps (see \cref{fig:ates}). If the physician is only interested in seeing the most important features TSR (\cref{fig:Ismail}) can be applied.  With the help of interpretability, a physician can gather evidence for the predicted classification and gain trust in the system (\cite{amann_explainability_2020}).
%\Cref{fig:Vis_univariate}  and \Cref{fig:Vis_multivariate} visualizes the interpretation  obtained with TSInterpret on NUN-CF (\Cref{fig:Nun-Cf}), CoMTE (\Cref{fig:ates}), LEFTIST (\Cref{fig:Leftist}), and TSR (\Cref{fig:Ismail}) for a 1D-conv ResNet on a data point from the UCR/UEA archive. 
In general, the visualization for the counterfactual approaches CoMTE and NUN-CF have a similar style, although NUN-CF visualizes univariate and CoMTE multivariate data. Note that CoMTE only visualizes the changed features. For the feature attribution methods LEFTIST and TSR, the visualizations have similar styles but a different color map. This is necessary as LEFTIST returns a positive and negative influence (range $[-1,1]$), while TSR returns normalized time slices and feature importances (range $[0,1]$).\par
The easy-to-use interface of TSInterpret allows the out-of-box usage of interpretability models on time series data for different use cases, time series types, and classification models. Further, the framework can be easily extended by inheriting from either \texttt{InterpretabilityBase} or one of the more customized classes (e.g., \texttt{Feature Attribution} or \texttt{InstanceBased}). All classes below the Output layer  also come with a default plot function to match the visualizations obtained by the already implemented algorithms.
%\todo[inline]{@Cedric: Move section below to conclusion?}
%Currently, TSInterpret omits ante-hoc interpretability. However, interpretability on recurrent neural networks primarily rely on ante-hoc mechanisms like attention mechanisms for model interpretability which include interpretability in the model design (\cite{rojat_explainable_2021}). Therefore, for now, no interpretability mechanisms tailored to recurrent neural networks are implemented. If model-agnostic interpretability algorithms for recurrent neural networks become available, the framework can be easily extended.
\section{Outlook}
TSInterpret provides a cross-backend unified API for the interpretability of time series classifications enabling interpretation generation with just three lines of code. In the first phase of the development, TSInterpret provides four interpretability algorithms for both uni- and multivariate time series classification and functions for visualizing the interpretation results. 
However, the Framework is easily extensible by inheritance and provides a simple API design.
In the future, we will include support for additional interpretability algorithms, time series prediction, as well as metrics to progress into an easy-to-use benchmarking tool for time series interpretability. In order to focus the research and application of time series prediction models not only on performance but also on understanding the model in a practical environment, a further development step is to integrate TSInterpret into the landscape of existing frameworks and thus establish it as an easily usable standard.

\section*{Acknowledgments}{This work was carried out with the support of the German Federal Ministry of Education and Research (BMBF) withhin the project "MetaLearn" (Grant 02P20A013).}

%%% Angabe der .bib-Datei (ohne Endung) / State .bib file (for BibTeX usage)
\bibliography{library} %\printbibliography if you use biblatex/Biber
\end{document}